%% file: main.tex
\PassOptionsToPackage{dvipsnames}{xcolor}
\documentclass{article}
\usepackage[preprint]{tmlr/tmlr}

\input{packages}
\input{macros}

\input{format}

\title{Modular Duality in Deep Learning}

\author{\name Jeremy Bernstein \email jbernstein@mit.edu \\
\name Laker Newhouse \email lakern@mit.edu \\
      \addr MIT CSAIL
}

\begin{document}

\maketitle

\input{section/00-abstract}
\input{section/01-intro}
\input{section/08-related-work}
\input{section/02-preliminaries}
\clearpage
\input{section/03-steepest}

\clearpage
\input{section/06-computational}

\input{section/09-discussion}

\section*{Acknowledgements}
Many ideas in this paper were developed jointly with Tim Large before he left to work at a tech company.
We are grateful to Phillip Isola for invaluable discussions. We also thank Jack Gallagher, Keller Jordan, Simo Ryu, Rogier Brussee, Tongzhou Wang, Victor Butoi, Jeffrey Cider and Volkan Cevher for helpful conversations.

\bibliography{refs}
\bibliographystyle{tmlr/tmlr}

\newpage
\clearpage

\end{document}

%% file: packages.tex
\usepackage[utf8]{inputenc}
\usepackage[T1]{fontenc}

\usepackage[dvipsnames]{xcolor}
\usepackage[nopatch=footnote]{microtype}
\usepackage{fontawesome}
\usepackage{dsfont}
\usepackage{verbatim}
\usepackage{pifont}

\usepackage{quoting}
\usepackage{titlesec}

\usepackage[pagebackref=true, hidelinks]{hyperref}
\usepackage{natbib}
\usepackage{url}

\usepackage[final,pdftex]{graphicx}
\usepackage{float}
\usepackage{caption}
\usepackage{tcolorbox}
\usepackage{booktabs,array}
\usepackage{tabularx}
\usepackage{makecell}
\usepackage{multirow}
\usepackage{tikz}
\usetikzlibrary{shapes.geometric,positioning}
\usepackage{pgfornament}

\usepackage[frozencache,cachedir=minted-output]{minted}
\usepackage[boxed]{algorithm2e}
\usepackage[noend]{algcompatible}

\usepackage{amsfonts,amssymb,amsmath,amsthm}
\usepackage[noabbrev,nameinlink,capitalise,compress]{cleveref}
\usepackage{mathtools}
\usepackage{bm}
\usepackage{mathabx}
\usepackage{thmtools,thm-restate}
\usepackage{resizegather}
\usepackage{nicefrac}
\usepackage{xfrac}

\usepackage{enumitem}
\usepackage{setspace}
\usepackage{regexpatch}

%% file: macros.tex
\newcommand{\for}{\mathsf{.forward}}
\newcommand{\mass}{\mathsf{.mass}}
\newcommand{\nor}{\mathsf{.norm}}

\newcommand{\lip}{\mathsf{.sensitivity}}

\newcommand{\dua}{\mathsf{.dualize}}

\newcommand{\module}{\mathsf{M}}
\newcommand{\atom}{\mathsf{A}}
\newcommand{\bond}{\mathsf{B}}

\newcommand{\relu}{\mathsf{ReLU}}

\newcommand{\linear}{\mathsf{Linear}}

\newcommand{\conv}{\mathsf{Conv2D}}

\newcommand{\embed}{\mathsf{Embed}}

\DeclareMathOperator*{\argmax}{arg\,max}
\DeclareMathOperator*{\argmin}{arg\,min}

\newcommand{\defeq}{\vcentcolon=}
\newcommand{\el}{\mathcal{L}}
\newcommand{\weights}{\mathcal{W}}
\newcommand{\inputs}{\mathcal{X}}
\newcommand{\outputs}{\mathcal{Y}}

\newcommand{\dualize}{\operatorname{dualize}}

\newcommand{\rms}{\mathrm{RMS}}
\newcommand{\out}{\mathrm{out}}
\newcommand{\inn}{\mathrm{in}}

\renewcommand{\phi}{\varphi}

\newcommand{\R}{\mathbb{R}}

\newcommand{\sign}{\operatorname{sign}}
\newcommand{\flatten}{\operatorname{flatten}}

\newcommand{\norm}[1]{\Vert {#1} \Vert}

\def\vg{{\bm{g}}}

\def\vt{{\bm{t}}}

\def\vv{{\bm{v}}}
\def\vw{{\bm{w}}}
\def\vx{{\bm{x}}}
\def\vy{{\bm{y}}}

\def\mG{{\bm{G}}}

\def\mM{{\bm{M}}}

\def\mT{{\bm{T}}}
\def\mU{{\bm{U}}}
\def\mV{{\bm{V}}}
\def\mW{{\bm{W}}}
\def\mX{{\bm{X}}}

\def\mSigma{{\bm{\Sigma}}}

%% file: format.tex
\renewcommand*{\backrefalt}[4]{\ifcase #1 No citations. \or Cited on page #2. \else Cited on pages #2. \fi}

\newcommand{\captiontitle}[1]{\textsf{\textbf{#1}}}

  \newtheorem{myexample}{Example}

  \newtheorem{myproposition}{Proposition}

  \newtheorem{mydefinition}{Definition}

\crefname{mydefinition}{Definition}{Definitions}

\SetAlCapSkip{1em}
\SetKwComment{Comment}{\small$\#$ }{}

%% file: section/00-abstract.tex
\begin{abstract}
An old idea in optimization theory says that since the gradient is a \textit{dual vector} it may not be subtracted from the weights without first being mapped to the \textit{primal space} where the weights reside. We take this idea seriously in this paper and construct such a duality map for general neural networks. Our map, which we call \textit{modular dualization}, forms a unifying theoretical basis for training algorithms that are a) \textit{fast} and b) \textit{scalable}. Modular dualization involves first assigning operator norms to layers based on the semantics of each layer, and then using these layerwise norms to recursively induce a duality map on the weight space of the full neural architecture. We conclude by deriving GPU-friendly algorithms for dualizing $\embed$, $\linear$ and $\conv$ layers---the latter two methods are based on a rectangular Newton-Schulz iteration \citep{kovarik1970iterative,bjoerck1971}. A variant of our methods was used to set speed records for training NanoGPT. Overall, we hope that our theory of modular duality will yield a next generation of fast and scalable optimizers for general neural architectures.
\end{abstract}

%% file: section/01-intro.tex
\section{Introduction}

In this paper, we pursue a rigorous and first-principles theoretical framework for designing neural network training algorithms. We hope that building such a framework will facilitate the design of a next generation of fast and scalable optimizers that are automatically tailored to different neural architectures.

While gradient descent is the workhorse of modern machine learning, the most vanilla form of the algorithm does not, in our view, pass a basic \textit{type check}. For a gradient update to type check, we insist that the gradient must be passed through a duality map before being multiplied by a learning rate and applied to the weights:
\begin{align}
    &\mathtt{weight} - \mathtt{LR}\,\mathtt{*}\,\mathtt{weight.grad} &&\texttt{type check: \textcolor{BrickRed}{failed!}}\\
    &\mathtt{weight} - \mathtt{LR}\,\mathtt{*}\,\mathtt{dualize}(\mathtt{weight.grad}) &&\texttt{type check: \textcolor{ForestGreen}{passed!}}
\end{align}
Why? The reason is that the loss function may not be equally smooth in all directions in weight space, and there is no reason for the sizes of different components of the raw gradient vector to respect this heterogeneity. In other words, \textit{the geometry of the loss function may be non-isotropic}. Insisting on a type check should force the user to become cognizant of this issue and to find a suitable duality map. A good duality map should adjust the size and direction of the gradient to respect the smoothness structure of the loss function.

Duality maps on vector spaces are commonplace in physics and applied math. Examples include the \textit{musical isomorphism} in differential geometry \citep{grosse2022metrics}, \textit{raising and lowering indices} in general relativity \citep{carroll2019spacetime} and the \textit{bra-ket notation} in quantum mechanics \citep{sakurai2020modern}. Duality maps are also central to several optimization theories including \textit{mirror descent} \citep{nemirovsky_yudin_1983}, \textit{natural gradient descent} \citep{amari2016information} and \textit{steepest descent on a normed space} \citep{Boyd_Vandenberghe_2004}. Despite the efforts of some prescient papers \citep{spectral-descent-4,flynn2017duality}, the latter kind of duality map involving normed vector spaces is yet to puncture the deep learning mainstream.

We believe that duality is a key theoretical concept that will help in building performant large-scale machine learning systems. To support this belief, we show in this paper that two important and seemingly disparate methods in contemporary optimization research may be seen as approximations to a single duality map. These methods are \textit{maximal update parameterization} \citep{Yang2021TensorPI}, which is aimed at scalable training, and \textit{Shampoo} \citep{Shi2023DistributedShampoo}, which is targeted at fast training. We show in \cref{sec:atomic-duality} that both methods emerge as partial approximations to a single duality map induced by the RMS--RMS operator norm.

The main contribution of this paper is to describe a procedure for constructing duality maps for general neural architectures. The procedure, which we call \textit{modular dualization}, works in three steps:
\begin{itemize}[leftmargin=1.45cm, itemindent=0cm]
    \item[\textbf{Step 1:}] Operator norms are assigned to individual layers based on the input-output semantics of each layer;
    \item[\textbf{Step 2:}] Based on these operator norms, duality maps are constructed for individual layers;
    \item[\textbf{Step 3:}] Given the layerwise duality maps and the structure of the neural architecture, a single duality map is recursively induced on the full weight space of the architecture. 
\end{itemize}

To instantiate this procedure for a rich family of neural architectures---including convolutional networks and transformers---we write down duality maps for $\linear$, $\embed$ and $\conv$ layers. We also provide GPU-friendly algorithms for computing these duality maps. Overall, we hope that modular dualization will help in the principled design of the machine learning systems of the future.

%% file: section/08-related-work.tex
\section{Related Work}

This paper constructs a duality map for general neural architectures. Our approach is based on assigning operator norms to individual network layers and using these layerwise norms to recursively induce a duality map on the full neural architecture. The most closely related prior work is a series of papers on \textit{spectral descent} \citep{spectral-descent-2,spectral-descent-4,spectral-descent-1} and a paper on \textit{duality structure gradient descent}  \citep{flynn2017duality}.

Spectral descent has been applied to restricted Boltzmann machines \citep{spectral-descent-2} and discrete graphical models \citep{spectral-descent-1}, but let us focus on the more closely related paper on spectral descent for deep learning \citep{spectral-descent-4}. In that paper, the authors propose assigning the Schatten-$\infty$ norm (a.k.a.\ spectral norm) to individual linear layers. This assignment is based on the observation that neural networks admit natural majorization bounds in the Schatten-$\infty$ norm. The authors call the corresponding duality map for linear layers the ``\#-operator''---a name presumably inspired by the musical isomorphism \citep{grosse2022metrics}. The authors propose a cheap approximation to the \#-operator based on sketching \citep{Martinsson_Tropp_2020}, and they also propose a way to mix RMSprop-style pre-conditioning information \citep{tieleman_rmsprop_2012} into the weight updates. In contrast to our work, the authors only derive duality maps for single linear layers, and these maps are then heuristically extended to all-layer updates. Nonetheless, the authors achieve substantial wall clock speedups using variants of spectral descent to train small networks.

Now, let us turn our attention to duality structure gradient descent \citep{flynn2017duality}, which constructs a duality map on the full weight space of the neural architecture based on identifying a \textit{Finsler structure} \citep{deimling1985nonlinear} inherent to neural networks. Similar to modular dualization, \citet{flynn2017duality}'s duality map works by assigning duality maps to each layer and then inducing a duality map on the full weight space. The substantial difference to our approach is that \citet{flynn2017duality} leverages a weighted sum ($L_1$ combination) of layerwise norms to construct his full duality map. This leads to optimization methods that only update a single layer at each iteration, and the methods need to be heuristically extended to achieve all-layer updates. In contrast, we leverage the modular norm \citep{modula}, which takes a weighted max ($L_\infty$ combination) of layerwise norms. In turn, our duality map leads directly to more conventional 
all-layer optimizers.

Another important difference between our work on modular duality and prior work on duality structure gradient descent is that we fully ``modularize'' our theory---meaning that our construction is explicitly recursive---and as such it is easy to code up into a software package. In this regard, we are inspired by a line of work that attempts to build optimization algorithms that automatically adapt to the structure of general computation graphs. The earliest work we know of in this category is the PhD thesis of \citet{Grant2004} on disciplined convex programming, which aims to infer the convexity properties of general functions by breaking them up into subexpressions and applying composition theorems from convex analysis. More recent progress in this vein includes work on universal majorization-minimization algorithms \citep{Streeter2022AutomaticallyBT,streeter2023universal} and related papers on automatic majorization \citep{Tran2015FastDT,agd-2023}.

%% file: section/02-preliminaries.tex
\section{Theoretical Preliminaries}

In this section, we introduce duality maps, a means of constructing duality maps based on norms, and finally a norm called the \textit{modular norm} that is well-suited to describe the geometry of general neural architectures.

\subsection{Duality Maps}\label{sec:steepest-descent}

Given a vector space $\mathcal{V}$, we say that a function $f:\mathcal{V} \to \R$ is a \textit{linear functional} on $\mathcal{V}$ if $f$ is linear. We define the \textit{dual space} $\mathcal{V}^*$ to be the set of linear functionals on the vector space $\mathcal{V}$. The dual space is itself a vector space provided that addition is defined pointwise $(f+g)(x) \defeq f(x) + g(x)$ and scalar multiplication is defined pointwise $(\alpha f)(x) \defeq \alpha f(x)$ for any scalar $\alpha$. By \textit{duality map} we simply mean any function that sends members of the dual vector space $\mathcal{V}^*$ to the primal vector space $\mathcal{V}$. The function need not be an involution.

Let $\el:\weights \to \R$ denote the loss of a differentiable machine learning model with weight space $\weights=\R^n$. The Taylor expansion of the loss at weight setting $\vw \in \weights$ is given by:
\begin{equation}\label{eq:taylor}
    \el(\vw+\Delta\vw) = \el(\vw) + \nabla_\vw\el(\vw)^\top \Delta \vw + \text{higher-order terms}.
\end{equation}
Observe that, in the first-order term, the gradient $\nabla_\vw\el(\vw)$ is acting as a linear functional: it is pairing with the weight vector $\Delta \vw \in \weights$ in a linear way to produce a real number. As such, we shall say that the gradient belongs to the dual weight space: $\nabla_\vw\el(\vw)\in\weights^*$. We shall forbid ourselves from directly subtracting a member of the dual weight space $\weights^*$ from the weight space $\weights$. If we would like to conduct a gradient descent update, then we had better find a duality map to send the gradient back to the primal space $\weights$.

This restriction may seem absurd! After all, here the weight space $\weights$ and its dual $\weights^*$ are both just $\R^n$. However, insisting upon this type check serves to remind us that the curvature of the loss function may be highly heterogeneous. The next section will show one way to construct duality maps to account for this.

\subsection{Steepest Descent on a Normed Space}\label{sec:steepest-descent-prelim}

Suppose that we have found a \textit{norm} $\norm{\cdot}:\weights\to\R$ and a \textit{sharpness parameter} $\lambda >0$ that serve as a good model of the higher-order terms in the Taylor expansion of the loss function given in \cref{eq:taylor}:
\begin{equation}\label{eq:taylor-approx}
    \el(\vw+\Delta\vw) \approx \el(\vw) + \nabla_\vw \el(\vw)^\top\Delta\vw + \frac{\lambda}{2}\cdot \norm{\Delta \vw}^2.
\end{equation}In other words, the norm provides a good characterization of the heterogeneity in curvature of the loss function. Then it makes sense to solve for a weight update $\Delta \vw$ by minimizing the right-hand side of \cref{eq:taylor-approx}. We will show that the minimizer can be expressed in terms of a \textit{dual norm} and a \textit{duality map}:

\begin{mydefinition}[Dual norm] Given a norm $\norm{\cdot}:\R^n\to\R$, the dual norm $\norm{\cdot}^\dagger$ of a vector $\vg \in \R^n$ is given by:
\begin{equation}
\norm{\vg}^\dagger\defeq\max_{\vt\in\R^n:\norm{\vt}=1} \vg^\top \vt.
\end{equation}
\end{mydefinition}

\begin{mydefinition}[Duality map based on a norm] Given a norm $\norm{\cdot}:\R^n\to\R$, we consider the duality map:
\begin{equation}
\dualize_{\norm{\cdot}} \vg \defeq \argmax_{\vt\in\R^n:\norm{\vt}=1} \vg^\top \vt,
\end{equation}
where, if the $\argmax$ is not unique, $\dualize_{\norm{\cdot}}$ returns any maximizer.    
\end{mydefinition}

Given these definitions, minimizing the expression in the right-hand side of \cref{eq:taylor-approx} can be done using the following standard proposition, for which \citet{bernstein2024old} provide a proof:
\begin{myproposition}[Steepest descent under a norm]\label{prop:steepest} For any $\vg \in \R^n$ thought of as ``the gradient'', any $\lambda \geq 0$ thought of as ``the sharpness'', and any norm $\norm{\cdot}:\R^n\to\R$ with dual norm $\norm{\cdot}^\dagger$ and duality map $\dualize_{\norm{\cdot}}$:
\begin{align}\label{eq:dual-steepest}
    \argmin_{\Delta \vw \in \R^n} \left[\vg^\top \Delta \vw + \frac{\lambda}{2} \, \norm{\Delta \vw}^2 \right] = - \frac{\norm{\vg}^\dagger}{\lambda} \times \dualize_{\norm{\cdot}} \vg.
\end{align}
\end{myproposition}
In words: to find the minimizer of a linear term penalized by a squared norm, we need only evaluate the dual norm and a duality map. In this paper, we focus on constructing a duality map for the \textit{modular norm}, which is defined on general neural architectures. The next section reviews duality maps for more standard norms.

\subsection{Basic Norms and Duality Maps}\label{sec:basic-norms}
Many basic norms and duality maps are already covered in prior work \citep{spectral-descent-1,spectral-descent-2,spectral-descent-4,flynn2017duality}. For some warmup examples, the following duality maps for vector norms are standard:
\begin{myexample}[Duality map for the Euclidean norm] For a vector $\vg\in\R^d$, we have $\dualize_{\norm{\cdot}_2} \vg = \vg / \norm{\vg}_2$.
\end{myexample}

\begin{myexample}[Duality map for the infinity norm] For a vector $\vg \in \R^d$, we have $\smash{\dualize_{\norm{\cdot}_\infty} \vg = \sign(\vg)}$, where the sign function is applied entrywise and we are free to take $\sign(0)=0$.
\end{myexample}
In neural networks, the weight spaces of individual layers tend to have matrix structure. And layers with the same shape weight matrix may have semantically different input and output spaces---think \textit{embedding} versus \textit{linear} layers in a transformer. As such, we will need duality maps for different \textit{induced operator norms}:
\begin{mydefinition}[Induced 
operator norm]\label{def:induced} Given a matrix $\mM\in\R^{d_\out \times d_\inn}$ and two normed vector spaces $(\R^{d_\inn},\norm{\cdot}_\alpha)$ and $(\R^{d_\out},\norm{\cdot}_\beta)$, the ``$\alpha$ to $\beta$'' induced operator norm is given by:
\begin{equation}
    \norm{\mM}_{\alpha\to\beta} = \max_{\substack{\vx \in \R^{d_\inn}}} 
    \frac{\norm{\mM\vx}_\beta}{\norm{\vx}_\alpha}.
\end{equation}
\end{mydefinition}
For tensors, we define the duality map via $\dualize_{\norm{\cdot}} \mG \defeq \argmax_{\norm{\mT}=1} \flatten(\mG)^\top \flatten(\mT)$. For linear layers, we will need the duality map for the $\rms\to\rms$ induced operator norm. This ends up as a rescaled version of the spectral norm duality map from prior work \citep{spectral-descent-4,flynn2017duality}.
\begin{myexample}[Duality map for the $\rms \to \rms$ operator norm] For a vector $\vv\in\R^d$, we define the RMS norm to be the normalized Euclidean norm: $\norm{\vv}_\rms = \norm{\vv}_2 / \sqrt{d}$. Given a matrix $\mW \in \R^{d_\out \times d_\inn}$, the $\rms \to \rms$ induced operator norm resolves to a rescaled spectral norm: $\norm{\mW}_{\rms \to \rms} = \sqrt{d_\inn / d_\out} \times \norm{\mW}_* $, where $\norm{\cdot}_*$ denotes the standard spectral norm. For a matrix $\mG \in \R^{d_\out \times d_\inn}$ with reduced singular value decomposition $\mG = \mU\mSigma\mV^\top$, the corresponding duality map is given by $\dualize_{\norm{\cdot}_{\rms \to \rms}} \mG = \sqrt{d_\out/d_\inn} \times \mU \mV^\top.$
\end{myexample}
And for embedding layers, we will need the duality map for the $\ell_1 \to \rms$ operator norm:
\begin{myexample}[Duality map for the $\ell_1 \to \rms$ operator norm] Given a matrix $\mW \in \R^{d_\out \times d_\inn}$, the $\ell_1 \to \rms$ induced operator norm resolves to the max $\rms$ norm of the columns: $\norm{\mW}_{\ell_1\to\rms} = \max_i \norm{\mathrm{col}_i(\mW)}_\rms$. For a matrix $\mG \in \R^{d_\out \times d_\inn}$, the corresponding duality map $\dualize_{\norm{\cdot}_{\ell_1 \to \rms}} \mG$ simply normalizes each column of $\mG$ to have unit RMS norm: $\mathrm{col}_i(\mG) \mapsto \mathrm{col}_i(\mG) / \norm{\mathrm{col}_i(\mG)}_\rms$ for each $i=1,...,d_\inn$.
\end{myexample}

\subsection{The Modular Norm}
\label{sec:modular-norm}

The \textit{modular norm} \citep{modula} is intended to help characterize the heterogeneous curvature of general neural architectures. The construction first defines an abstract \textit{module} type along with a notion of what is a good, or \textit{well-normed}, module. Then \textit{combination rules} are given for constructing new well-normed modules from a library of existing well-normed modules. So modules are a special case of \textit{combinator pattern} from functional programming \citep{haskellwiki_combinator}. Modules are also related to a \textit{monoidal category} from category theory \citep{fong2019invitation}. We begin by defining the abstract notion of a \textit{module}:

\begin{mydefinition}[Module]\label{def:module} Given input vector space $\inputs$, output vector space $\outputs$ and weight vector space $\weights$, a module $\module$ is an object with the following four attributes:
\begin{enumerate}[label=\normalfont(\alph*)]
\setlength\itemsep{0em}
    \item a function, $\module\for: \weights\times\inputs\to\outputs$, which maps an input and a weight vector to an output;
    \item a number, $\module\mass\geq0$, which is used to set the proportion of feature learning that this module contributes to any supermodule;
    \item a number, $\module\lip \geq0$, which estimates the module's sensitivity to input perturbations;
    \item a norm over the weight space, $\module\nor : \weights \to \R_{\geq0}$, sometimes abbreviated to just $\norm{\cdot}_\module$.
\end{enumerate}
\end{mydefinition}
We shall care most about modules that are \textit{well-normed}, which amounts to requiring that the forward function is Lipschitz-continuous in the weights with constant 1 and in the inputs with constant $\module\lip$:

\begin{mydefinition}[Well-normed module]\label{def:well-normed}
Let $\module$ be a module on $(\inputs, \outputs, \weights)$, where the input and output spaces have respective norms $\norm{\cdot}_{\inputs}$ and $\norm{\cdot}_{\outputs}$. $\module$ is well-normed if for all inputs $\vx\in\inputs$ and weights $\vw \in\weights$:
\begin{align}
\norm{\nabla_\vw\module\for(\vw,\vx)\diamond\Delta \vw}_\outputs &\leq \module\nor(\Delta \vw) &&\text{for all }\Delta\vw\in\weights;\\
\norm{\nabla_\vx\module\for(\vw,\vx)\diamond\Delta \vx}_\outputs & \leq \module\lip * \norm{\Delta \vx}_\inputs &&\text{for all }\Delta\vx\in\inputs.
\end{align}
\end{mydefinition}
The $\diamond$ operator denotes summation over any shared tensor indices. This definition of well-normed-ness can be used as a guiding principle in the design of a library of atomic (i.e. handwritten) modules. First, norms should be assigned to the input and output space of each module based on the semantics of $\module\for$. Then a norm $\module\nor$ should be assigned to the module's weight space and a number $\module\lip$ should be chosen to make the module well-normed. Examples are given in \cref{sec:atomic-duality}.

Given such a library of well-normed atomic modules, a compound module built through any arbitrary sequence of \textit{module compositions} and \textit{module concatenations} is automatically well-normed \citep{modula}. And if the atomic modules in the library are not only well-normed but are also \textit{smooth} in an appropriate sense, then \citet{modula} give an automatic procedure for computing \textit{sharpness coefficients} for any compound module built from the library. The relevant definition of module composition is as follows:

\begin{mydefinition}[Module composition]\label{def:composition} Consider module $\module_1$ with input, output and weight space $(\inputs_1,\outputs_1,\weights_1)$ and module $\module_2$ with input, output and weight space $(\inputs_2,\outputs_2,\weights_2)$. $\module_1$ and $\module_2$ are \textit{composable} if $\inputs_2 = \outputs_1$. Their composite module $\module=\module_2\circ\module_1$ has input, output and weight space $(\inputs_1,\outputs_2,\weights_1 \times \weights_2)$ and attributes:
\begin{enumerate}[label=\normalfont(\alph*)]
\setlength\itemsep{0em}
\item $\module\for((\vw_1,\vw_2),\vx)) = \module_2\for(\vw_2,\module_1\for(\vw_1,\vx))$;%

\item $\module\mass = \module_1\mass + \module_2\mass$;%

\item $\module\lip = \module_1\lip * \module_2\lip$;%

\item $\module\nor((\vw_1, \vw_2))$ given by:
\begin{equation*}\label{eq:norm_composition}
     \max\left(
    \module_2\lip * \frac{\module\mass}{\module_1\mass} * \module_1\nor(\vw_1),
    \frac{\module\mass}{\module_2\mass} * \module_2\nor(\vw_2)\right),
\end{equation*}
where if $\module_1\mass$ or $\module_2\mass$ is zero, the corresponding term in the $\max$ is set to zero.
\end{enumerate}
\end{mydefinition}
So the composite norm is taken to be a weighted max over the norms of the two sub-modules, where the weight space of the first module is coupled to the input sensitivity of the second module. The module masses provide freedom to tune the importance of each sub-module in the norm, and \citet{modula} prove that module mass provides precise control over the amount of feature learning that can happen in each sub-module.

Module concatenation is defined in a similar way to module composition:
\begin{mydefinition}[Module concatenation]\label{def:concatenation} Consider module $\module_1$ with input, output and weight space $(\inputs_1,\outputs_1,\weights_1)$ and module $\module_2$ with input, output and weight space $(\inputs_2,\outputs_2,\weights_2)$. We say that $\module_1$ and $\module_2$ are concatenatable if their input spaces match: $\inputs_1 = \inputs_2$. The tuple module $\module=(\module_1,\module_2)$ has input, output and weight space $(\inputs_1, \outputs_1\times\outputs_2, \weights_1\times\weights_2)$ and the following list of attributes:
\begin{enumerate}[label=\normalfont(\alph*)]
\setlength\itemsep{0em}
\item $\module\for((\vw_1,\vw_2),\vx)) = (\module_1\for(\vw_1,\vx), \module_2\for(\vw_2,\vx))$;%
\item $\module\mass = \module_1\mass + \module_2\mass$;%
\item $\module\lip = \module_1\lip + \module_2\lip$;%
\item $\module\nor(\vw_1, \vw_2)$ given by:
\begin{equation*}\label{eq:norm_addition}
    \max\left(
    \frac{\module\mass}{\module_1\mass} * \module_1\nor(\vw_1),
    \frac{\module\mass}{\module_2\mass} * \module_2\nor(\vw_2)\right),
\end{equation*}
where if $\module_1\mass$ or $\module_2\mass$ is zero, the corresponding term in the $\max$ is set to zero.
\end{enumerate}
\end{mydefinition}

A shortcoming of the paper by \citet{modula} is that the power of the modular norm is not fully leveraged. In particular, the authors do \textit{modular normalization} of training, where weight updates to modules are sometimes just na\"ively divided by their norm. In this paper we make fuller use of the geometry implied by the modular norm by constructing the corresponding duality map, which we call \textit{modular dualization}.

%% file: section/03-steepest.tex
\setlist{topsep=0pt,itemsep=0pt,parsep=5pt,partopsep=0pt}

\section{Modular Dualization}
\label{sec:steepest-descent-in-the-modular-norm}

In this section, we construct a duality map for general neural architectures. Our strategy is to first write down duality maps for atomic modules, i.e.\ individual layers. We then extend to arbitrary compound modules, i.e.\ full neural networks, by showing how duality maps should pass through composition and concatenation.

\subsection{Duality Maps for Atomic Modules}
\label{sec:atomic-duality}

To construct a duality map for an atomic module $\atom$, the idea is to first fix norms on the input and output spaces that respect the semantics of $\atom\for$. We should select norms that describe both how large we would like the inputs and outputs to be, and in what geometry we would like the outputs to evolve. Then we place a norm on the weight space such that $\atom$ is well-normed: this is typically the operator norm (\cref{def:induced}) induced by the input and output norms. Finally we are in position to solve for the duality map, which we shall call $\atom\dua$. We now give some examples of this procedure for the basic layer types of $\linear$, $\embed$ and $\conv$. The results are summarized in \cref{tab:dualize}.

We start with the canonical example of an atomic module:
\begin{myexample}[The $\linear$ module] The $\linear$ module sends inputs from $\inputs = \R^{d_\inn}$ to outputs in $\outputs = \R^{d_\out}$. The weight space is given by the matrix space $\weights = \R^{d_\out \times d_\inn}$. We endow the $\linear$ module with attributes:

\begin{enumerate}
    \setlength\itemsep{0em}
    \item $\linear\for(\mW,\vx) = \mW\vx$, the matrix-vector product;
    \item $\linear\lip = 1$;
    \item $\linear\mass = \mu$, where $\mu\geq 0$ is a hyperparameter;
    \item $\linear\nor(\mW) = \norm{\mW}_{\rms\to\rms}$, the $\rms\to\rms$ induced operator norm.
\end{enumerate}
Since the $\linear$ module is intended to map to and from vectors of roughly unit $\rms$ norm, we place the $\rms$ norm on both the input and output space: $\norm{\cdot}_\inputs = \norm{\cdot}_\rms$ and $\norm{\cdot}_\outputs = \norm{\cdot}_\rms$. Then $\linear$ is well-normed if the inputs and weights belong to the unit balls $\left\{\vx\in\R^{d_\inn}:\norm{\vx}_\inputs \leq 1\right\}$ and $\left\{\mW \in \R^{d_\out\times d_\inn}:\linear\nor(\mW)\leq1\right\}$. Referring back to \cref{sec:basic-norms}, the duality map corresponding to $\linear\nor$ is then given by:

\begin{enumerate}
\setcounter{enumi}{4}
\setlength\itemsep{0em}
    \item $\linear\dua(\mG) = \sqrt{\frac{d_\out}{d_\inn}} \times \mU\mV^\top$, where the gradient $\mG \in \R^{d_\out \times d_\inn}$ has reduced SVD $\mG = \mU \mSigma \mV^\top$.
\end{enumerate}
\end{myexample}

This single duality map recovers essential features of both \textit{maximal update parameterization} \citep[$\mu$P]{Yang2021TensorPI} and \textit{Shampoo} \citep{Gupta2018ShampooPS}. In particular, the factor of $\sqrt{d_\out / d_\inn}$ in $\linear\dua$ recovers spectral update scaling \citep{my-spectral} that leads to $\mu$P. (Initializing such that $\linear\nor(\mW)=1$ also recovers $\mu$P initialization scaling.) And the mapping $\mG \mapsto \mU\mV^\top$ is equivalent to Shampoo without accumulation \citep{bernstein2024old}. As such, we believe that duality maps may help reconcile different strands of deep learning research and provide a unifying basis for fast and scalable training algorithms.

The $\embed$ module provides a useful counterpoint to the $\linear$ module. The difference between the two modules stems from the fact that the input spaces of $\embed$ and $\linear$ have different semantics.
\begin{myexample}[The $\embed$ module] The $\embed$ module sends inputs from $\inputs = \R^{d_\inn}$ to outputs in $\outputs = \R^{d_\out}$. The weight space is given by the matrix space $\weights = \R^{d_\out \times d_\inn}$. We endow the $\embed$ module with attributes:

\begin{enumerate}
\setlength\itemsep{0em}
    \item $\embed\for(\mW,\vx) = \mW\vx$, the matrix-vector product;
    \item $\embed\lip = 1$;
    \item $\embed\mass = \mu$, where $\mu\geq 0$ is a hyperparameter;
    \item $\embed\nor(\mW) = \norm{\mW}_{\ell_1\to\rms}$, the $\ell_1\to\rms$ induced operator norm.
\end{enumerate}
$\embed$ is intended to map from one-hot vectors to vectors of roughly unit $\rms$ norm, so we place the $\ell_1$ norm on the input space and the $\rms$ norm on the output space: $\norm{\cdot}_\inputs = \norm{\cdot}_{\ell_1}$ and $\norm{\cdot}_\outputs = \norm{\cdot}_\rms$. Then $\embed$ is well-normed if the inputs and weights belong to the unit balls $\left\{\vx\in\R^{d_\inn}:\norm{\vx}_\inputs \leq 1\right\}$ and $\left\{\mW \in \R^{d_\out\times d_\inn}:\embed\nor(\mW)\leq1\right\}$. Referring back to \cref{sec:basic-norms}, the duality map for $\embed\nor$ is:

\begin{enumerate}
\setcounter{enumi}{4}
\setlength\itemsep{0em}
    \item $\embed\dua(\mG)$ performs the mapping $\mathrm{col}_j(\mG) \mapsto \frac{\mathrm{col}_j(\mG)}{ \norm{\mathrm{col}_j(\mG)}_\rms}$ for each column index $j=1,...,d_\inn$.
\end{enumerate}
\end{myexample}

Finally, we consider a $\conv$ module with a $k\times k$ kernel. $\conv$ has a more involved tensor structure than $\linear$ and $\embed$. The calculations work by slicing up the weight tensor into a collection of $k^2$ matrices.

\begin{myexample}[The $\conv$ module] The $\conv$ module sends inputs from $\inputs = \R^{W_\inn\times H_\inn\times d_\inn}$ to outputs in $\outputs = \R^{W_\out \times H_\out \times d_\out}$. We think of this as mapping an input image of width $W_\inn$, height $H_\inn$ and with $d_\inn$ color channels to an output image of width $W_\out$, height $H_\out$ and with $d_\out$ color channels. The weight space is given by the tensor space $\weights = \R^{d_\out \times d_\inn\times k \times k}$, where $k$ is the kernel size. We endow $\conv$ with attributes:

\begin{enumerate}
\setlength\itemsep{0em}
    \item $\conv\for(\mW,\vx) = \mW\circledast\vx$, where $\circledast$ denotes 2D convolution;
    \item $\conv\lip = 1$;
    \item $\conv\mass = \mu$, where $\mu\geq 0$ is a hyperparameter;
    \item $\conv\nor(\mW) = k^2 \max_{i,j=1}^k \norm{\mW_{\cdot \cdot ij}}_{\rms \to \rms}$, the max $\rms\to\rms$ norm over kernel indices.
\end{enumerate}
We would like pixel intensities in the inputs and outputs to be order one and undergo order one change. We formalize this by taking the input and output norms to be the spatial maximum of the RMS norms of all the color channel vectors: $\norm{\vx}_\inputs = \max_{w=1}^{W_\inn}\max_{h=1}^{H_\inn} \norm{\vx_{wh\cdot}}_\rms$ and $\norm{\vy}_\outputs = \max_{w=1}^{W_\out}\max_{h=1}^{H_\out} \norm{\vy_{wh\cdot}}_\rms$. Then $\conv$ is well-normed if the inputs and weights belong to the unit balls $\left\{\vx\in\R^{W_\inn\times H_\inn\times d_\inn}:\norm{\vx}_\inputs \leq 1\right\}$ and $\left\{\mW \in \R^{d_\out\times d_\inn\times k \times k}:\conv\nor(\mW)\leq1\right\}$. Since the duality map for a max of norms decouples into one duality map per sub-norm, the duality map corresponding to $\conv\nor$ is given by:

\begin{enumerate}
\setcounter{enumi}{4}
\setlength\itemsep{0em}
    \item $\conv\dua(\mG)$ does $\mG_{\cdot\cdot ij} \mapsto \frac{1}{k^2} \sqrt{\frac{d_\out}{d_\inn}} \times \mU_{ij} \mV_{ij}^\top$, where $\mG_{\cdot\cdot ij}$ has reduced SVD $\mU_{ij}\mSigma_{ij}\mV_{ij}^\top$.
\end{enumerate}
\end{myexample}

\input{figure/dualize}

\subsection{Duality Maps for Bond Modules}

\citet{modula} define another class of basic modules: \textit{bond modules}. Bonds are handwritten modules without weights. An example of a bond is the $\relu$ nonlinearity. For a bond $\bond$, the weight space is the zero vector space $\weights =\{0\}$ and the modular norm $\bond\nor = 0\mapsto 0$. As such, the corresponding duality map is also $\bond\dua = 0\mapsto 0$. In a software package, one need not write norms or duality maps for bond modules.

\subsection{Duality Maps for Compound Modules}
\label{sec:compound-duality}

First, given two composable modules $\module_1$ and $\module_2$, the duality map for the composite $\module= \module_2\circ\module_1$ is given by:
\begin{equation}\label{eq:composite-duality}
    \module\dua(\vg_1, \vg_2) = \left(\frac{1}{\module_2\lip} * \frac{\module_1\mass}{\module\mass} * \module_1\dua(\vg_1), \frac{\module_2\mass}{\module\mass} * \module_2\dua(\vg_2)\right).
\end{equation}
And second, given two concatenatable modules $\module_1$ and $\module_2$, the duality map for the tuple $\module = (\module_1, \module_2)$ is:
\begin{equation}\label{eq:tuple-duality}
    \module\dua(\vg_1, \vg_2) = \left(\frac{\module_1\mass}{\module\mass} * \module_1\dua(\vg_1), \frac{\module_2\mass}{\module\mass} * \module_2\dua(\vg_2)\right).
\end{equation}
The proofs of \cref{eq:composite-duality,eq:tuple-duality} follow in a straightforward manner from
\cref{def:composition,def:concatenation}.

%% file: figure/dualize.tex
\newcommand{\stackargs}[2]{\renewcommand{\arraystretch}{1.1}\begin{tabular}{@{}c@{}}#1 \\ #2\end{tabular}\renewcommand{\arraystretch}{1.4}}

\renewcommand{\stackargs}[2]{#1 #2}

\renewcommand{\arraystretch}{1.4}
\begin{table}
    \centering
    \begin{tabular}{@{}cccc@{}}
        \toprule
         \textsf{\textbf{Module}} & \textsf{\textbf{Weight Space}} $\bm\weights$ & $\bm{\mathsf{Module}\nor}$ & $\bm{\mathsf{Module}\dua}$ \\
        \midrule
        $\linear$ & $\R^{d_\out \times d_\inn}$ & $\mW \mapsto\norm{\mW}_{\rms \to \rms}$ & $\mG \mapsto \sqrt{\frac{d_\out}{d_\inn}} \times \mU \mV^\top$ \\
        
        $\embed$ & $\R^{d_\out \times d_\inn}$ & $\mW \mapsto\norm{\mW}_{\ell_1 \to \rms}$ & $\mathrm{col}_j(\mG) \mapsto \frac{\mathrm{col}_j(\mG)}{ \norm{\mathrm{col}_j(\mG)}_\rms}$ \\
        
        $\conv$ & $\R^{d_\out \times d_\inn \times k \times k}$ & $\mW \mapsto k^2 \max_{i,j=1}^k \norm{\mW_{\cdot \cdot ij}}_{\rms \to \rms}$ & $\mG_{\cdot\cdot ij} \mapsto \frac{1}{k^2} \sqrt{\frac{d_\out}{d_\inn}} \times \mU_{ij} \mV_{ij}^\top$ \\
        \bottomrule
    \end{tabular}
    \caption{\captiontitle{Duality maps for three atomic modules: $\bm\linear$, $\bm\embed$, and $\bm\conv$.} These atomic modules are sufficient to build convolutional neural networks and transformers. In $\linear\dua$, we let $\mU \mSigma \mV^\top$ denote the reduced SVD of the gradient matrix $\mG$. In $\conv\dua$, we let $\smash{\mU_{ij} \mSigma_{ij} \mV_{ij}^\top}$ denote the reduced SVD of the slice of the gradient tensor $\mG_{\cdot\cdot ij}$ at kernel indices $i$ and $j$. \cref{sec:fast} provides GPU-friendly algorithms for computing these duality maps based on a family of Newton-Schulz iterations.}
    \label{tab:dualize}
\end{table}
\renewcommand{\arraystretch}{1}

%% file: section/06-computational.tex
\section{Fast Duality Maps}
\label{sec:fast}

For modular dualization to be practically feasible, we need ways of computing duality maps quickly. Inspecting the duality maps listed in \cref{tab:dualize}, we see that $\embed\dua$ is easy to implement since it just involves computing vector norms of matrix columns. But $\linear\dua$ and $\conv\dua$ involve the projection:
\begin{equation}\label{eq:polarize}
    \mG = \mU\mSigma \mV^\top \mapsto \mU\mV^\top,
\end{equation}
where $\mU\mSigma \mV^\top$ is the reduced SVD of the matrix $\mG$. Since computing SVDs can be slow \citep{spectral-descent-4,flynn2017duality}, here we discuss three fast approximations to \cref{eq:polarize} via sketching, iterations for inverse matrix roots, and a family of \textit{rectangular Newton-Schulz} iterations. Which method works best in practice may depend on the condition number of the matrix $\mG$ or the available computational resources.

\subsection{Sketching}

Sketching is a randomized method \citep{Martinsson_Tropp_2020} that can be used to build low-rank approximations to the SVD. \citet{spectral-descent-4} already used sketching to provide a fast approximation to their $\#$-operator. More recent papers have experimented with sketching in the context of Shampoo-type algorithms \citep{sketchy}. A potential downside of approximating \cref{eq:polarize} via sketching is that randomized SVD methods usually try to accurately approximate the largest singular values of a matrix \citep[Section 11.2]{Martinsson_Tropp_2020} while the value of \cref{eq:polarize} may lie in its action on the small singular values.

\subsection{Iterations for Inverse Matrix Roots}

Given a full rank matrix $\mG$ with reduced SVD $\mU\mSigma\mV^\top$, we have that:
\begin{equation}
    \mU\mV^\top = (\mG\mG^\top)^{-\sfrac{1}{4}} \,\mG\, (\mG^\top\mG)^{-\sfrac{1}{4}} = (\mG\mG^\top)^{-\sfrac{1}{2}} \,\mG = \mG\, (\mG^\top\mG)^{-\sfrac{1}{2}}.
\end{equation}
This provides a route to approximating \cref{eq:polarize} since one can compute inverse matrix roots such as $(\mG\mG^\top)^{-\sfrac{1}{2}}$ via Newton iteration \citep{lakic}. This is discussed in Chapter 7 of \citet{higham}'s book and also see \citet{Anil2020ScalableSecondOrder}'s paper. Care must be taken with inverses whenever the matrix $\mG$ is ill-conditioned.

\subsection{Rectangular Newton-Schulz Iteration}

We developed a ``rectangular Newton-Schulz iteration'' for computing $\mU\mV^\top$ by adapting Equation 5.22 in \citet{higham}'s book for computing the ``matrix sign function''. We later discovered that this iteration has a long history \citep{kovarik1970iterative, bjoerck1971}. In short, the method works by first normalizing the matrix $\mG$ according to $\mX_0 = \mG / \norm{\mG}_{\ell_2 \to \ell_2}$ (or alternatively $\mX_0 = \mG / \norm{\mG}_F$) and then iterating:
    \begin{equation}\label{eq:newton-schulz}
        \mX_{t+1} = \frac{3}{2} \cdot \mX_t - \frac{1}{2} \cdot \mX_t \mX_t^\top \mX_t,
    \end{equation}
    then as $t\to\infty$, the sequence $\mX_t \to \mU \mV^\top$. To see this, one can plot the univariate cubic function $f(x) \defeq \tfrac{3}{2} \cdot x - \tfrac{1}{2}\cdot x^3$ and see that, for $0 < x < \sqrt{3}$, iterating this cubic will push $x$ closer and closer to $+1$. The final step is to realize that the effect of the iteration in \cref{eq:newton-schulz} is to apply this cubic $f(x)$ to each singular value of $\mX_t$. This shows that the spectral normalization $\mX_0 = \mG / \norm{\mG}_{\ell_2 \to \ell_2}$ is stronger than what is required: we need only ensure that $\mX_0$ has singular values no greater than $\sqrt{3}$ for the iteration to converge.

    The iteration in \cref{eq:newton-schulz} has the advantage over sketching that it always works on all singular values, and since the iteration does not compute inverse matrix roots it is well-behaved even on low-rank matrices.

    Finally, there are in fact a family of degree $2n+1$ polynomial iterations of the form
    \begin{equation}
    \mX_{t+1} = a \cdot \mX_t + b  \cdot \mX_t \mX_t^\top \mX_t + c \cdot (\mX_t \mX_t^\top)^2 \mX_t + ... + z \cdot (\mX_t \mX_t^\top)^n \mX_t
    \end{equation}
    for suitable $a,b,c,...,z$  that could be used instead of \cref{eq:newton-schulz}. One should choose coefficients $a, b,c,...,z$ so that the univariate polynomial $g(x) = a \cdot x + b \cdot x^3 + c \cdot x^5 + ... + z\cdot x^{2n+1}$ is a suitable approximation to $\sign(x)$. One may try to further accelerate the iteration by ``tuning'' the coefficients $a,b,c,...,z$ empirically.

%% file: section/09-discussion.tex
\clearpage
\section{Discussion}
\label{sec:epilogue}

This paper develops the theory of \textit{modular duality} and the procedure of \textit{modular dualization} as means to construct duality maps for general neural architectures. Here, we comment on implications and connections.

\subsection{A Type System for Deep Learning}

Part of the inspiration for this work is the idea of building a fully-fledged \textit{type system} for deep learning. We think that activation spaces should be typed by their intended norm and the intended size of activations in that norm. This information would help in the construction of well-normed modules (see $\cref{sec:atomic-duality}$). Modules should be typed according to \cref{def:module}. And, as suggested in the introduction, gradients should be explicitly typed as dual vectors. A duality map should flip the type of a dual vector to a primal vector. We plan to use the Modula deep learning package \citep{modula} as a testbed for these ideas.

\subsection{Neural Network Speedrunning}

We believe that the ideas in this paper can help in the design of faster training methods. In fact, a new NanoGPT training speed record was recently set \citep{jordan2024cifar10} using a Newton-Schulz-based duality map. We communicated the method to Keller Jordan through our workshop paper \citep{bernstein2024old}.

\subsection{Modular Duality: A Unifying Theoretical Framework for Fast and Scalable Training}

An important topic in contemporary optimization research is the design of fast and scalable training methods for neural networks. In fact, the theme of the Optimization for Machine Learning workshop at this year's NeurIPS conference is ``scaling up optimization'' \citep{opt2024}. Two popular methods in this research space are \textit{maximal update parameterization} \citep[$\mu$P]{Yang2021TensorPI}, which allows for increasing network width without changing the optimal learning rate, and \textit{Shampoo} \citep{Gupta2018ShampooPS}, a variant of which \citep{Shi2023DistributedShampoo} won a speed challenge at the inaugural AlgoPerf optimization competition \citep{Dahl2023AlgoPerf}.

We showed in \cref{sec:atomic-duality} that essential features of both $\mu$P and Shampoo are recovered from the single duality map $\linear\dua$. We think that, on a basic theoretical level, $\mu$P and Shampoo should be viewed as partial approximations to this duality map. This observation helps put $\mu$P and Shampoo on a consistent theoretical footing, orients the methods with respect to overlooked prior work on spectral descent \citep{spectral-descent-4} and duality structure gradient descent \citep{flynn2017duality}, and suggests new ways to generalize these methods to arbitrary layer types and network architectures via the modular norm and modular dualization.

\subsection{On the Alignment of Activations and Updates}

Recent work \citep{my-spectral,everett2024scaling,modula} has singled out the following question as important to the design of scalable deep learning systems: \textit{to what extent do gradient updates to neural network layers align with incoming activation vectors?} This question is important since it helps inform how large weight updates need to be to induce a certain amount of change in layer outputs. Duality maps such as $\linear\dua$ and $\conv\dua$ may help simplify the answer to this question, since they project gradients to scaled semi-orthogonal matrices for which all singular values have the same magnitude.

\subsection{A Numerical Paradox: \textit{The Weights Don't Change!}}

Past work \citep{NEURIPS2019_0d1a9651,math9182246} has pointed out an apparent paradox in deep learning: the weights seem to move a vanishing amount from initialization in the limit of large network width. This finding has motivated a substantial amount of work on linearized training dynamics \citep{NTKjacot}. We attempted to resolve this paradox in prior work by showing that the weights move a roughly constant amount at any width when the change is measured in spectral norm \citep{my-spectral}. But duality maps change the story again: $\linear\dua$ ramps up the stable rank of updates, so the weights should move a non-trivial relative amount at large width \textit{even in the Frobenius norm}---provided the batch size is not too small.

\section{Conclusion}

This paper has proposed a recursive procedure called \textit{modular dualization} for building duality maps for general neural architectures. The procedure unifies past strands of optimization research on Shampoo \citep{Gupta2018ShampooPS} and $\mu$P \citep{Yang2021TensorPI}. Partial implementations have already led to significant wall-clock speedups in transformer training \citep{jordan2024cifar10}. The rectangular Newton-Schulz iteration provides a GPU-friendly and numerically stable means of dualizing under the $\rms\to\rms$ operator norm, while avoiding some of the downsides of sketching-based approaches \citep{spectral-descent-4}. Overall, we hope that our theory of \textit{modular duality} provides a clarifying toolkit for the design and analysis of deep learning systems.

%% file: main.bbl
\begin{thebibliography}{39}
\providecommand{\natexlab}[1]{#1}
\providecommand{\url}[1]{\texttt{#1}}
\expandafter\ifx\csname urlstyle\endcsname\relax
  \providecommand{\doi}[1]{doi: #1}\else
  \providecommand{\doi}{doi: \begingroup \urlstyle{rm}\Url}\fi

\bibitem[Amari(2016)]{amari2016information}
Shun-ichi Amari.
\newblock \emph{Information Geometry and Its Applications}.
\newblock Springer, 2016.

\bibitem[Anil et~al.(2020)Anil, Gupta, Koren, Regan, and Singer]{Anil2020ScalableSecondOrder}
Rohan Anil, Vineet Gupta, Tomer Koren, Kevin Regan, and Yoram Singer.
\newblock Scalable second order optimization for deep learning.
\newblock \emph{arXiv:2002.09018}, 2020.

\bibitem[Bernstein \& Newhouse(2024)Bernstein and Newhouse]{bernstein2024old}
Jeremy Bernstein and Laker Newhouse.
\newblock Old optimizer, new norm: An anthology.
\newblock In \emph{Workshop on Optimization for Machine Learning}, 2024.

\bibitem[Bernstein et~al.(2023)Bernstein, Mingard, Huang, Azizan, and Yue]{agd-2023}
Jeremy Bernstein, Chris Mingard, Kevin Huang, Navid Azizan, and Yisong Yue.
\newblock {A}utomatic {G}radient {D}escent: {D}eep {L}earning without {H}yperparameters.
\newblock \emph{arXiv:2304.05187}, 2023.

\bibitem[Bj\"{o}rck \& Bowie(1971)Bj\"{o}rck and Bowie]{bjoerck1971}
\r{A}ke Bj\"{o}rck and C.~Bowie.
\newblock An iterative algorithm for computing the best estimate of an orthogonal matrix.
\newblock \emph{SIAM Journal on Numerical Analysis}, 1971.

\bibitem[Boyd \& Vandenberghe(2004)Boyd and Vandenberghe]{Boyd_Vandenberghe_2004}
Stephen Boyd and Lieven Vandenberghe.
\newblock \emph{Convex Optimization}.
\newblock Cambridge University Press, 2004.

\bibitem[Carlson et~al.(2015{\natexlab{a}})Carlson, Cevher, and Carin]{spectral-descent-2}
David Carlson, Volkan Cevher, and Lawrence Carin.
\newblock Stochastic spectral descent for restricted {B}oltzmann machines.
\newblock In \emph{International Conference on Artificial Intelligence and Statistics}, 2015{\natexlab{a}}.

\bibitem[Carlson et~al.(2016)Carlson, Hsieh, Collins, Carin, and Cevher]{spectral-descent-1}
David Carlson, Ya-Ping Hsieh, Edo Collins, Lawrence Carin, and Volkan Cevher.
\newblock Stochastic spectral descent for discrete graphical models.
\newblock \emph{Selected Topics in Signal Processing}, 2016.

\bibitem[Carlson et~al.(2015{\natexlab{b}})Carlson, Collins, Hsieh, Carin, and Cevher]{spectral-descent-4}
David~E. Carlson, Edo Collins, Ya-Ping Hsieh, Lawrence Carin, and Volkan Cevher.
\newblock Preconditioned spectral descent for deep learning.
\newblock In \emph{Neural Information Processing Systems}, 2015{\natexlab{b}}.

\bibitem[Carroll(2019)]{carroll2019spacetime}
Sean~M. Carroll.
\newblock \emph{Spacetime and Geometry: An Introduction to General Relativity}.
\newblock Cambridge University Press, 2019.

\bibitem[Dahl et~al.(2023)Dahl, Schneider, Nado, Agarwal, Sastry, Hennig, Medapati, Eschenhagen, Kasimbeg, Suo, Bae, Gilmer, Peirson, Khan, Anil, Rabbat, Krishnan, Snider, Amid, Chen, Maddison, Vasudev, Badura, Garg, and Mattson]{Dahl2023AlgoPerf}
George~E. Dahl, Frank Schneider, Zachary Nado, Naman Agarwal, Chandramouli~Shama Sastry, Philipp Hennig, Sourabh Medapati, Runa Eschenhagen, Priya Kasimbeg, Daniel Suo, Juhan Bae, Justin Gilmer, Abel~L. Peirson, Bilal Khan, Rohan Anil, Mike Rabbat, Shankar Krishnan, Daniel Snider, Ehsan Amid, Kongtao Chen, Chris~J. Maddison, Rakshith Vasudev, Michal Badura, Ankush Garg, and Peter Mattson.
\newblock Benchmarking neural network training algorithms.
\newblock \emph{arXiv:2306.07179}, 2023.

\bibitem[Deimling(1985)]{deimling1985nonlinear}
Klaus Deimling.
\newblock \emph{Nonlinear Functional Analysis}.
\newblock Springer Berlin, Heidelberg, 1985.

\bibitem[Everett et~al.(2024)Everett, Xiao, Wortsman, Alemi, Novak, Liu, Gur, Sohl-Dickstein, Kaelbling, Lee, and Pennington]{everett2024scaling}
Katie~E. Everett, Lechao Xiao, Mitchell Wortsman, Alexander~A. Alemi, Roman Novak, Peter~J. Liu, Izzeddin Gur, Jascha Sohl-Dickstein, Leslie~Pack Kaelbling, Jaehoon Lee, and Jeffrey Pennington.
\newblock Scaling exponents across parameterizations and optimizers.
\newblock In \emph{International Conference on Machine Learning}, 2024.

\bibitem[Feinberg et~al.(2023)Feinberg, Chen, Sun, Anil, and Hazan]{sketchy}
Vladimir Feinberg, Xinyi Chen, Y.~Jennifer Sun, Rohan Anil, and Elad Hazan.
\newblock Sketchy: Memory-efficient adaptive regularization with frequent directions.
\newblock In \emph{Neural Information Processing Systems}, 2023.

\bibitem[Flynn(2017)]{flynn2017duality}
Thomas Flynn.
\newblock The duality structure gradient descent algorithm: Analysis and applications to neural networks.
\newblock \emph{arXiv:1708.00523}, 2017.

\bibitem[Fong \& Spivak(2019)Fong and Spivak]{fong2019invitation}
Brendan Fong and David~I. Spivak.
\newblock \emph{An Invitation to Applied Category Theory: Seven Sketches in Compositionality}.
\newblock Cambridge University Press, 2019.

\bibitem[Grant(2004)]{Grant2004}
Michael~Charles Grant.
\newblock \emph{Disciplined Convex Programming}.
\newblock {PhD} dissertation, Stanford University, 2004.

\bibitem[Grosse(2022)]{grosse2022metrics}
Roger Grosse.
\newblock Metrics.
\newblock Lecture 3 of CSC2541: Neural Net Training Dynamics, 2022.

\bibitem[Gupta et~al.(2018)Gupta, Koren, and Singer]{Gupta2018ShampooPS}
Vineet Gupta, Tomer Koren, and Yoram Singer.
\newblock Shampoo: Preconditioned stochastic tensor optimization.
\newblock In \emph{International Conference on Machine Learning}, 2018.

\bibitem[{Haskell Wiki Contributors}(2007)]{haskellwiki_combinator}
{Haskell Wiki Contributors}.
\newblock Combinator pattern.
\newblock Haskell Wiki, 2007.
\newblock URL \url{https://wiki.haskell.org/Combinator_pattern}.

\bibitem[Higham(2008)]{higham}
Nicholas~J. Higham.
\newblock \emph{Functions of Matrices}.
\newblock Society for Industrial and Applied Mathematics, 2008.

\bibitem[Jacot et~al.(2018)Jacot, Gabriel, and Hongler]{NTKjacot}
Arthur Jacot, Franck Gabriel, and Clement Hongler.
\newblock Neural tangent kernel: {C}onvergence and generalization in neural networks.
\newblock In \emph{Neural Information Processing Systems}, 2018.

\bibitem[Jesus et~al.(2021)Jesus, Antunes, da~Costa, Dorogovtsev, Mendes, and Aguiar]{math9182246}
Ricardo~J. Jesus, Mário~L. Antunes, Rui~A. da~Costa, Sergey~N. Dorogovtsev, José F.~F. Mendes, and Rui~L. Aguiar.
\newblock Effect of initial configuration of weights on training and function of artificial neural networks.
\newblock \emph{Mathematics}, 2021.

\bibitem[Jordan(2024)]{jordan2024cifar10}
Keller Jordan.
\newblock New training speed record for @karpathy’s 124{M}-parameter {NanoGPT} setup: 3.28 {F}ineweb validation loss in 3.7{B} training tokens.
\newblock \url{https://x.com/kellerjordan0/status/1842300916864844014}, 2024.

\bibitem[Kovarik(1970)]{kovarik1970iterative}
Zdislav Kovarik.
\newblock Some iterative methods for improving orthonormality.
\newblock \emph{SIAM Journal on Numerical Analysis}, 1970.

\bibitem[Lakić(1998)]{lakic}
Slobodan Lakić.
\newblock On the computation of the matrix k-th root.
\newblock \emph{Journal of Applied Mathematics and Mechanics}, 1998.

\bibitem[Large et~al.(2024)Large, Liu, Huh, Bahng, Isola, and Bernstein]{modula}
Tim Large, Yang Liu, Minyoung Huh, Hyojin Bahng, Phillip Isola, and Jeremy Bernstein.
\newblock Scalable optimization in the modular norm.
\newblock In \emph{Neural Information Processing Systems}, 2024.

\bibitem[Lee et~al.(2019)Lee, Xiao, Schoenholz, Bahri, Novak, Sohl-Dickstein, and Pennington]{NEURIPS2019_0d1a9651}
Jaehoon Lee, Lechao Xiao, Samuel Schoenholz, Yasaman Bahri, Roman Novak, Jascha Sohl-Dickstein, and Jeffrey Pennington.
\newblock Wide neural networks of any depth evolve as linear models under gradient descent.
\newblock In \emph{Neural Information Processing Systems}, 2019.

\bibitem[Martinsson \& Tropp(2020)Martinsson and Tropp]{Martinsson_Tropp_2020}
Per-Gunnar Martinsson and Joel~A. Tropp.
\newblock Randomized numerical linear algebra: Foundations and algorithms.
\newblock \emph{Acta Numerica}, 2020.

\bibitem[Nemirovsky \& Yudin(1983)Nemirovsky and Yudin]{nemirovsky_yudin_1983}
Arkady~S. Nemirovsky and David~B. Yudin.
\newblock \emph{Problem complexity and method efficiency in optimization}.
\newblock Wiley, 1983.

\bibitem[OPT(2024)]{opt2024}
OPT.
\newblock {Optimization for Machine Learning}, 2024.
\newblock URL \url{https://opt-ml.org/}.

\bibitem[Sakurai \& Napolitano(2020)Sakurai and Napolitano]{sakurai2020modern}
J.~J. Sakurai and Jim Napolitano.
\newblock \emph{Modern Quantum Mechanics}.
\newblock Cambridge University Press, 2020.

\bibitem[Shi et~al.(2023)Shi, Lee, Iwasaki, Gallego-Posada, Li, Rangadurai, Mudigere, and Rabbat]{Shi2023DistributedShampoo}
Hao-Jun~Michael Shi, Tsung-Hsien Lee, Shintaro Iwasaki, Jose Gallego-Posada, Zhijing Li, Kaushik Rangadurai, Dheevatsa Mudigere, and Michael Rabbat.
\newblock A distributed data-parallel {PyTorch} implementation of the distributed {S}hampoo optimizer for training neural networks at-scale.
\newblock \emph{arXiv:2309.06497}, 2023.

\bibitem[Streeter(2023)]{streeter2023universal}
Matthew Streeter.
\newblock Universal majorization-minimization algorithms.
\newblock \emph{arXiv:2308.00190}, 2023.

\bibitem[Streeter \& Dillon(2022)Streeter and Dillon]{Streeter2022AutomaticallyBT}
Matthew~J. Streeter and Joshua~V. Dillon.
\newblock Automatically bounding the {T}aylor remainder series: Tighter bounds and new applications.
\newblock \emph{arXiv:2212.11429}, 2022.

\bibitem[Tieleman \& Hinton(2012)Tieleman and Hinton]{tieleman_rmsprop_2012}
Tijmen Tieleman and Geoffrey Hinton.
\newblock {RMSprop}.
\newblock \emph{Coursera: Neural Networks for Machine Learning}, Lecture 6.5, 2012.

\bibitem[Tran et~al.(2015)Tran, Ono, and Vincent]{Tran2015FastDT}
Dung~T. Tran, Nobutaka Ono, and Emmanuel Vincent.
\newblock Fast {DNN} training based on auxiliary function technique.
\newblock \emph{International Conference on Acoustics, Speech and Signal Processing}, 2015.

\bibitem[Yang \& Hu(2021)Yang and Hu]{Yang2021TensorPI}
Greg Yang and Edward~J. Hu.
\newblock Tensor programs {IV}: Feature learning in infinite-width neural networks.
\newblock In \emph{International Conference on Machine Learning}, 2021.

\bibitem[Yang et~al.(2023)Yang, Simon, and Bernstein]{my-spectral}
Greg Yang, James~B. Simon, and Jeremy Bernstein.
\newblock A spectral condition for feature learning.
\newblock \emph{arXiv:2310.17813}, 2023.

\end{thebibliography}
